\documentclass{article}

\PassOptionsToPackage{numbers}{natbib}
\usepackage[main, final]{neurips_2026}
\usepackage{enumitem}

\usepackage[utf8]{inputenc} 
\usepackage[T1]{fontenc}    
\usepackage{hyperref}       
\usepackage{url}            
\usepackage{booktabs}       
\usepackage{amsfonts}       
\usepackage{nicefrac}       
\usepackage{microtype}      
\usepackage{xcolor}         
\usepackage{amsmath,amsthm}
\usepackage{multirow}
\usepackage{adjustbox}
\usepackage{lipsum} 
\usepackage{booktabs} 
\usepackage{soul}
\usepackage{algorithm}
\usepackage{algorithmic}
\usepackage{wrapfig} 
\usepackage{tikz}
\usepackage{colortbl}
\usepackage{fontawesome5}

\title{Why Are LLMs Confidently Wrong?
Correcting Overconfident Errors via Causal Head Intervention}

\author{%
  \textbf{Jing Ren\textsuperscript{1,\#}} \quad
  \textbf{Bowen Li\textsuperscript{1,\#}} \quad
  \textbf{Ziqi Xu\textsuperscript{1,*}} \quad
  \textbf{Xuechao Yang\textsuperscript{1}} \quad
  \textbf{Feng Xia\textsuperscript{1}} \\[4pt]
  \textsuperscript{1}RMIT University \\[5pt]
  {\ttfamily \{S4065838, S3890442\}@student.rmit.edu.au} \\
  {\ttfamily \{ziqi.xu, xuechao.yang, feng.xia\}@rmit.edu.au}
    \vspace{2mm}
  \\
  \textsuperscript{\#}Equal contribution \quad
  \textsuperscript{*}Corresponding author \\[6pt]
  \href{https://aisuko.github.io/confix}
  {\faGithub\ \texttt{https://aisuko.github.io/confix}}
}

\begin{document}

\maketitle

\begin{abstract}
Modern large language models (LLMs) often exhibit overconfident errors, assigning high confidence to incorrect predictions and thereby undermining reliability in real-world use. Existing calibration methods typically rely on post-hoc adjustment or retraining, operating at the output level without addressing the underlying causes of miscalibration. We propose \textbf{CoCHI} (\textbf{C}lass-\textbf{o}riented \textbf{C}ausal \textbf{H}ead \textbf{I}ntervention), a framework that improves calibration by identifying and intervening on internal components responsible for distinct confidence behaviours. By analysing predictions through correctness-confidence patterns, CoCHI enables targeted intervention on model internals, reducing overconfident errors while preserving correct predictions. Our approach provides a mechanistic perspective on calibration, linking reliability to identifiable structures within the model rather than treating it as an output-level issue. Experiments across multiple LLMs show consistent improvements in calibration metrics, including Expected Calibration Error, Negative Log-Likelihood, and Brier Score, while maintaining accuracy on five multiple-choice question answering benchmarks. These results suggest that effective calibration requires not only adjusting outputs, but also addressing the underlying causal mechanisms within the model. 
\end{abstract}

\section{Introduction}

Modern neural models have achieved remarkable performance across a wide range of tasks. However, they frequently produce overconfident errors, assigning high confidence to incorrect predictions~\cite{yadkori2024believe,steyvers2025large,ren2026spiking,NguyenXCHXZ26} and thereby undermining reliability in high-stakes applications. This issue is further amplified in large language models (LLMs)~\cite{achiam2023gpt,liu2024deepseek,grattafiori2024llama}, which have become a dominant paradigm in modern AI and are widely deployed in domains such as healthcare, education, and scientific reasoning, where incorrect yet confident outputs can lead to significant downstream risks~\cite{bang2025hallulens,ravichander2025halogen,wu2025combating,okutomi2025stable,YangXLMCSZR26}.

To address this problem, a large body of work has focused on \emph{calibration}, aiming to align predicted confidence with empirical accuracy~\cite{ulmer2024calibrating,huang2026cats,ren2026trust}. Existing approaches, such as temperature scaling and its variants, primarily operate at the output level by adjusting logits or confidence scores, while more recent methods introduce learned calibrators or inference-time interventions~\cite{guo2017calibration}. Despite their effectiveness in some settings, these approaches share a fundamental limitation: they treat miscalibration as a surface-level phenomenon and do not address the \emph{internal mechanisms} that give rise to overconfident errors. As a result, they often fail to generalise and provide limited insight into why models are confidently wrong.

In this work, we take a different perspective and ask: \emph{what internal components of a model are responsible for overconfident errors, and can we intervene on them directly?}  Recent advances in mechanistic interpretability~\cite{nam2025causal,vig2020investigating,meng2022locating,ChengXLLYLL26} have made significant progress in moving beyond correlational post-hoc explanations toward causally grounded, intervention-based understanding of how LLMs implement computations internally. In particular, causal intervention methods such as causal head gating~\cite{nam2025causal}, causal mediation analysis~\cite{vig2020investigating,XuCLLLW23}, and causal tracing~\cite{meng2022locating} allow us to identify internal components that are causally responsible for specific model behaviours by systematically perturbing internal activations and observing the resulting change in the model’s output distribution. However, existing interpretability approaches largely focus on post-hoc analysis and explanation, rather than leveraging these insights to actively improve model reliability. Consequently, the connection between internal computational circuits and miscalibration remains largely unexplored.

To bridge this gap, we propose \textbf{CoCHI} (\textbf{C}lass-\textbf{o}riented \textbf{C}ausal \textbf{H}ead \textbf{I}ntervention), a framework that improves calibration by directly targeting the internal causes of overconfident errors in large language models. Our key insight is that miscalibration is not merely an output-level issue, but arises from specific internal components that systematically promote incorrect yet highly confident predictions or introduce noise into otherwise correct ones. To capture this, we decompose model predictions into four calibration classes based on correctness and confidence, and perform class-conditioned causal analysis to identify attention heads associated with distinct behaviours. Based on this analysis, CoCHI constructs a structured intervention that selectively suppresses heads responsible for overconfident errors while removing non-contributing components in correct predictions. This results in a minimal, targeted modification to the model’s internal computation that improves calibration without altering model parameters, providing a mechanism-aware alternative to conventional post-hoc approaches.

\paragraph{Contributions.}
Our main contributions are as follows:
\begin{itemize}
\item We propose a structural causal mask for calibration, operating directly on model internals rather than relying on post-hoc output adjustment. 
\item We partition model predictions into four fine-grained classes based on correctness and confidence, enabling precise identification of attention heads that causally promote overconfident errors or fail to support correct predictions.
\item We introduce {CoCHI} (Class-oriented Causal Head Intervention), an inference-time framework that performs class-aware causal analysis and constructs a structured mask to selectively suppress problematic attention heads.
\item Extensive experiments across five multiple-choice question answering benchmarks show that CoCHI consistently improves calibration metrics while preserving accuracy, outperforming traditional post-hoc calibration methods.
\end{itemize}

\section{Related Work}
\paragraph{Uncertainty Estimation and Overconfident Errors.}
A large body of work studies uncertainty estimation in neural models, including Bayesian approaches~\cite{mackay1992bayesian,graves2011practical,blundell2015weight,hernandez2015probabilistic,welling2011bayesian}, deep ensembles~\cite{lakshminarayanan2017simple}, and post-hoc techniques such as temperature scaling~\cite{guo2017calibration}, which rescale logits using a learned temperature parameter. In the context of LLMs, uncertainty estimation and calibration have received increasing attention~\cite{xia2025influences}. Recent efforts extend post-hoc calibration to LLMs, including adaptive temperature scaling~\cite{xie2024calibrating} and learned calibrators such as Thermometer~\cite{shen2024thermometer}, aiming to better align predicted confidence with empirical accuracy. Other approaches adjust hidden representations, for example by training a linear layer on top of the final hidden states for confidence prediction~\cite{liu2024litcab}, or exploit prompt-based strategies and self-consistency to estimate uncertainty~\cite{tian2023just,zhang2024don,zhou2023navigating,mielke2022reducing,ren2026trust}. Despite their effectiveness in some settings, these methods primarily operate at the output or representation level, but recent work such as Inference-Time Intervention~\cite{li2023inference} shows that undesirable behaviours (e.g., hallucination) are governed by internal mechanisms. These methods primarily enhance truthful representations, while overlooking the role of mechanisms that specifically drive overconfident errors, where incorrect predictions are assigned high confidence scores. Recent work has shown that attention heads encode rich signals about model uncertainty, enabling unsupervised uncertainty estimation without additional training~\cite{vazhentsev2025uncertainty}. However, these approaches primarily focus on extracting uncertainty signals from internal representations, without modifying the model’s behaviour. This limitation motivates our approach, which targets the structural causal components responsible for overconfident errors and performs structured intervention to improve calibration.

\paragraph{Circuit Discovery in Mechanistic Interpretability}
Circuit discovery aims to identify the minimal subgraph of a language model responsible for a specific behaviour. The seminal work of \cite{wang2023interpretability} manually identified the Indirect Object Identification (IOI) circuit in GPT-2 Small, comprising 28 attention heads across seven functional classes. To scale this process, \citet{conmy2023towards} proposed Automatic Circuit Discovery (ACDC), an iterative pruning method based on activation patching, which successfully rediscovered known circuits such as Greater-Than and recovered most components of the IOI circuit. Subsequent approaches, including Efficient Attribution Patching (EAP), further improved the efficiency and scalability of automated circuit discovery~\cite{hsu2025efficient}. Beyond circuit discovery, mechanistic interpretability has increasingly incorporated tools from causal inference~\cite{pearl2009causality}. These approaches leverage counterfactual interventions and structured perturbations to uncover how structural heads contribute to model outputs~\cite{geiger2021causal,kaddour2025causal,wu2022causal,LiYXXSY26}. In particular, causal head gating identifies functional roles of attention heads by assigning graded scores for facilitation, interference, or irrelevance based on their causal effect on task performance, going beyond purely correlational analyses~\cite{nam2025causal}. Recent mechanistic studies have shown that verbalised confidence in LLMs is governed by dedicated internal circuits that are largely decoupled from factual correctness~\cite{zhao2026wired}. These findings suggest that overconfidence arises not from miscalibration at the output layer, but from specific internal mechanisms that consistently generate confident responses regardless of their accuracy~\cite{joshi2025calibration}. Building on this insight, we identify and suppress attention heads that contribute to high-confidence errors, leading to improved calibration and reliability.

\section{Preliminaries}

We briefly introduce the key components for identifying the causal roles of attention heads in transformer models via differentiable intervention.

\paragraph{Transformer with Head-wise Gating.}
Consider a transformer model with $L$ layers and $H$ attention heads per layer. For each attention head $(\ell, h)$, we introduce a continuous gating variable $G_{\ell,h} \in [0,1]$ that scales the output of the head:
\begin{equation}
Z_{\ell,h} = G_{\ell,h} \cdot \big( A_{\ell,h} V_{\ell,h} \big),
\end{equation}
where $A_{\ell,h}$ denotes the attention weights (after softmax) and $V_{\ell,h}$ denotes the value projections. 

The gated head outputs are concatenated and projected as in standard multi-head attention. By varying $G_{\ell,h}$, we perform soft interventions on individual heads without modifying model parameters.

\paragraph{Gate Optimisation.}
The gating variables are optimised while keeping model parameters frozen. Given a dataset $\mathcal{D}$, we minimise the negative log-likelihood objective:
\begin{equation}
\mathcal{L}(G) = - \mathbb{E}_{(x,y)\sim \mathcal{D}} \log P(y \mid x; G),
\end{equation}
where $P(y \mid x; G)$ denotes the model output under gating configuration $G$. 

Optionally, a regularisation term can be introduced to encourage sparsity or density of the gating values. This formulation allows us to assess the causal contribution of each head by observing how scaling its output affects model behaviour.

\paragraph{Dual Optimisation and Role Identification.}
To disambiguate causal importance from incidental effects, we optimise two sets of gates:
\begin{itemize}
    \item $G^{+}$: optimised under a retention bias, encouraging gates towards 1,
    \item $G^{-}$: optimised under a sparsity bias, encouraging gates towards 0.
\end{itemize}

The divergence between $G^{+}$ and $G^{-}$ reveals the causal role of each head: heads that remain active under sparsity are necessary for the task, while those suppressed even under retention bias are likely harmful.

\paragraph{Causal Head Taxonomy.}
Based on the optimised gates, each attention head is assigned a causal role. \textbf{Facilitating heads} contribute positively to performance and remain active under sparsity ($G^{-}$), indicating necessity. \textbf{Interfering heads} degrade performance and are suppressed even under retention bias ($G^{+}$), indicating harmful contributions. \textbf{Irrelevant heads} have negligible impact, with gating values varying under different regularisation regimes. This taxonomy provides a causal, rather than correlational, characterisation of model internals, validated through targeted ablation and intervention.

\section{Methodology}
\begin{figure}
    \centering
    \includegraphics[width=0.98\linewidth]{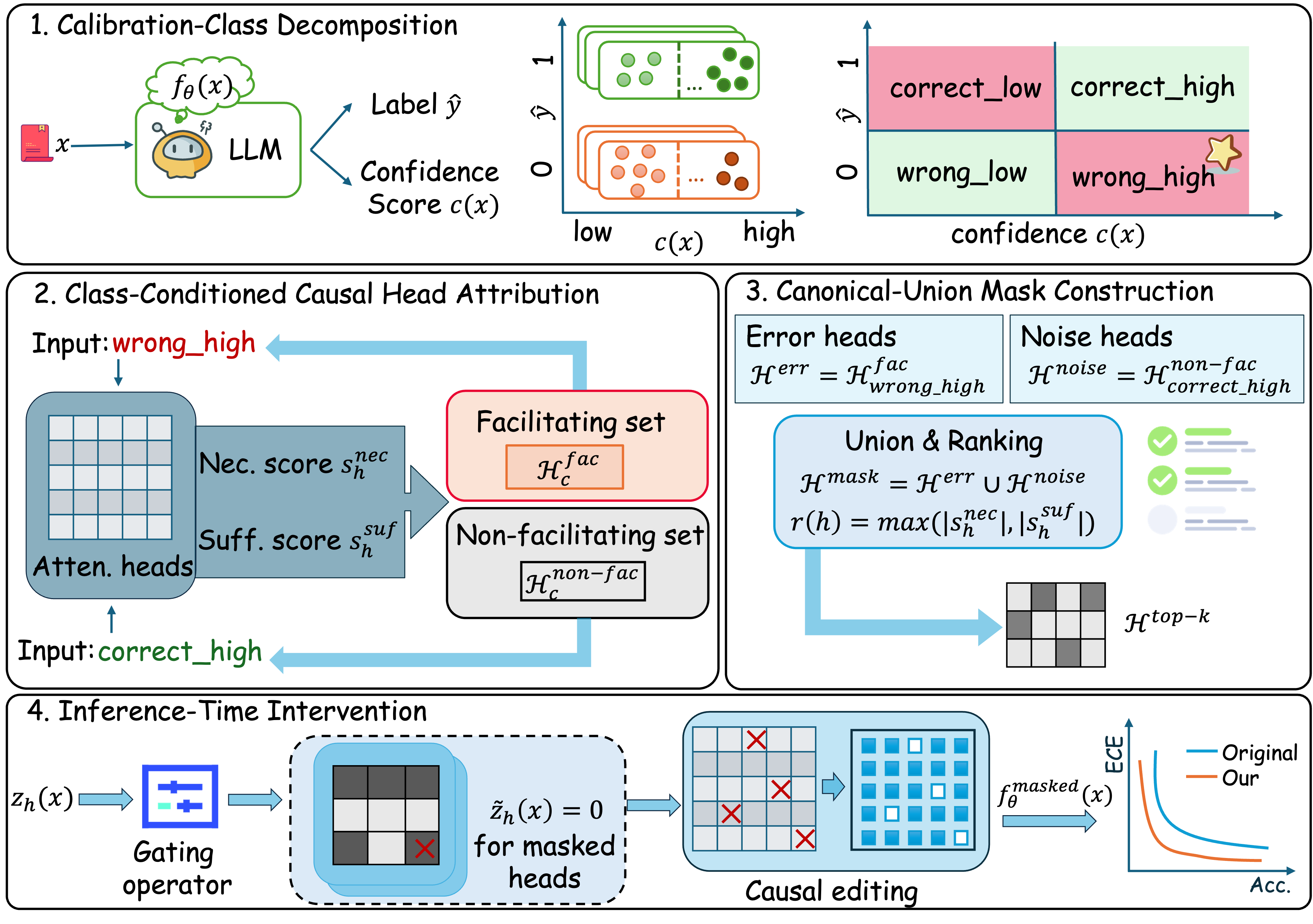}
    \caption{The framework of CoCHI.}
    \label{fig:fm}
\end{figure}

We reframe calibration as a problem of causal intervention on model heads. Specifically, our framework identifies and suppresses causal heads responsible for overconfident errors, while removing sources of noise in otherwise correct predictions. This perspective provides a mechanistic view of miscalibration, linking reliability to identifiable structural components within the model. An overview of \textbf{CoCHI} is shown in Figure~\ref{fig:fm}.

\subsection{Problem Setup}

Let $f_\theta(x)$ denote a pretrained language model that maps an input $x$ to a probability distribution $p_\theta(\cdot \mid x)$ over a fixed answer space $\mathcal{Y}$. The model prediction is defined as:
\begin{equation}
\hat{y}(x) = \arg\max_{y \in \mathcal{Y}} p_\theta(y \mid x).
\end{equation}

The corresponding confidence is defined as:
\begin{equation}
c(x) = \max_{y \in \mathcal{Y}} p_\theta(y \mid x).
\end{equation}

Let $y^*$ denote the ground-truth label. A prediction is correct if $\hat{y}(x) = y^*$. We focus on the failure mode of \emph{overconfident errors}, defined as incorrect predictions with high confidence:
\begin{equation}
\hat{y}(x) \neq y^*, \quad c(x) \ge \tau_c,
\end{equation}
where $\tau_c$ is a predefined confidence threshold.

Our goal is to design an inference-time intervention that reduces such errors by targeting their underlying internal causes, thereby improving calibration while preserving predictive accuracy.

\subsection{Calibration-Class Decomposition}

Rather than treating calibration as a global property of output probabilities, we decompose model behaviour into four calibration classes based on correctness and confidence:
\[
\mathcal{C} = \{ \texttt{correct\_high}, \texttt{correct\_low}, \texttt{wrong\_high}, \texttt{wrong\_low} \}.
\]
Formally, this defines a partition over predictions induced by correctness ($\hat{y}(x) = y^*$ or $\hat{y}(x) \neq y^*$) and confidence ($c(x) \ge \tau_c$ or $c(x) < \tau_c$). This decomposition isolates overconfident errors ($\texttt{wrong\_high}$) as a distinct subset, enabling us to analyse the structural causal head that give rise to miscalibrated confidence, rather than conflating them with other types of errors.

\subsection{Class-Conditioned Causal Head Attribution}

Following Causal Head Gating (CHG)~\cite{nam2025causal}, we attribute model behaviour to individual attention heads by estimating their causal contribution. Let $\mathcal{H}$ denote the set of all attention heads across $l$ layers. For each head $h \in \mathcal{H}$, CHG computes two scores with respect to a target behaviour defined on a dataset $\mathcal{D}$: a \emph{necessary score} $s_h^{\text{nec}}$, measuring the drop in performance when the head is ablated, and a \emph{sufficient score} $s_h^{\text{suf}}$, measuring how well the head alone can reproduce the behaviour when isolated.

In our setting, we apply CHG in a class-conditioned manner. For each calibration class $c \in \mathcal{C}$, we construct a subset $\mathcal{D}_c$ consisting of inputs whose predictions fall into class $c$, and perform CHG using the model’s own predictions as the target signal. This yields a \emph{necessary score} $s_h^{\text{nec}}(c)$ and a \emph{sufficient score} $ s_h^{\text{suf}}(c)$ for each head. The necessary score measures how strongly a head persists when the optimisation is encouraged to minimise the number of active heads, whereas the sufficient score measures how strongly a head remains selected when the optimisation is encouraged to retain all heads compatible with good task performance.

CoCHI is trained in three sequential stages. First, an unregularised optimisation stage ($\lambda = 0$) is performed to obtain a shared task-relevant initialisation:
\begin{equation}
z^{0}_{l,h} \xrightarrow{\text{stage 1}} \hat{z}_{l,h}.
\end{equation}

The resulting logits $\hat{z}_{l,h}$ are then copied into two separate optimisation branches:
\begin{equation}
\hat{z}_{l,h} \rightarrow z^{+}_{l,h}, \quad \hat{z}_{l,h} \rightarrow z^{-}_{l,h},
\end{equation}
corresponding to the positive-L1 and negative-L1 stages, respectively.

During each forward pass, the learned logit is converted into a multiplicative gate using a sigmoid function and applied directly to the attention head output:
\begin{equation}
\tilde{x}_{l,h} = \sigma(z_{l,h}) \cdot x_{l,h},
\end{equation}
where $x_{l,h} \in \mathbb{R}^{T \times d_h}$ denotes the output of head $h$ at layer $l$, and $T$ is the sequence length. A gate value close to $0$ suppresses the head, while a value close to $1$ preserves its contribution.

The optimisation objective is defined as:
\begin{equation}
\mathcal{L} =
-\frac{1}{|S|}\sum_{t \in S}
\log p_\theta(x_{t+1}\mid x_{\leq t})
-
\lambda \cdot \frac{1}{LH}
\sum_{l,h}
\mathrm{clamp}(z_{l,h},-4,4),
\end{equation}
where the first term is the next-token negative log-likelihood and the second term is the regularisation objective. The logits are clamped to $[-4,4]$ before regularisation to stabilise optimisation and encourage near-binary gate values.

Two complementary optimisation stages are then performed. In the positive-L1 stage ($\lambda = +0.1$), the optimiser is encouraged to suppress unnecessary heads, so only heads whose removal significantly increases the NLL retain large gate values. These heads correspond to necessary heads. In the negative-L1 stage ($\lambda = -0.1$), the optimiser instead encourages larger gate values, identifying heads that are sufficient to maintain low task loss.

At the end of optimisation, the final sigmoid gate values are recorded as:
\begin{equation}
s_h^{\text{nec}}(c) = \sigma(z^{+}_{l,h})^{(\mathrm{final})},
\end{equation}
\begin{equation}
s_h^{\text{suf}}(c) = \sigma(z^{-}_{l,h})^{(\mathrm{final})},
\end{equation}
which define the necessary and sufficient scores, respectively. 

A binary threshold $\tau = 0.5$ is then applied to derive discrete causal roles:
\begin{equation}
\mathrm{role}_{l,h} =
\begin{cases}
\textbf{facilitating}, & s_h^{\text{nec}}(c) \geq \tau \ \wedge\ s_h^{\text{suf}}(c) \geq \tau, \\
\textbf{interfering}, & s_h^{\text{nec}}(c) < \tau \ \wedge\ s_h^{\text{suf}}(c) \geq \tau, \\
\textbf{irrelevant}, & \text{otherwise}.
\end{cases}
\end{equation}

A head is considered \emph{facilitating} for class $c$ if both scores exceed a threshold $\tau$, indicating that it is both necessary and sufficient for producing the corresponding behaviour; otherwise, it is treated as non-facilitating, encompassing both interfering and irrelevant heads.

This procedure yields two sets per class, $\mathcal{H}_c^{\text{fac}}$ and $\mathcal{H}_c^{\text{non-fac}}$. Crucially, this class-conditioned decomposition associates specific structural causal head with distinct calibration behaviours and forms the basis for targeted intervention in subsequent steps.

\subsection{Canonical-Union Mask Construction}

We construct a targeted structural causal mask for calibration by combining two complementary types of heads. First, from the $\texttt{wrong\_high}$ class, we extract the facilitating heads:
\begin{equation}
\mathcal{H}^{\text{err}} = \mathcal{H}_{\texttt{wrong\_high}}^{\text{fac}},
\end{equation}
which contribute to generating high-confidence incorrect predictions. These heads can be interpreted as forming the targeted structural causal mask that amplify erroneous outputs.

Second, from the $\texttt{correct\_high}$ class, we extract the non-facilitating heads:
\begin{equation}
\mathcal{H}^{\text{noise}} = \mathcal{H}_{\texttt{correct\_high}}^{\text{non-fac}},
\end{equation}
which do not support correct predictions and may introduce variability in confidence estimation.

We then define the candidate mask set as the union $\mathcal{H}^{\text{mask}} = \mathcal{H}^{\text{err}} \cup \mathcal{H}^{\text{noise}}$.
To avoid excessive intervention, we rank heads by their attribution strength:
\begin{equation}
r(h) = \max\big(|s_h^{\text{nec}}|, |s_h^{\text{suf}}|\big),
\end{equation}
and select the top-$k$ heads to form a compact mask $\mathcal{H}^{\text{top-}k}$. This ranking focuses the intervention on the most influential components while preserving the overall functionality of the model. 


\subsection{Inference-Time Intervention}

Given the identified set of influential heads $\mathcal{H}^{\text{top-}k}$, we perform a structured intervention on the model’s internal computation at inference time. Rather than modifying model parameters or post-processing outputs, our approach directly alters the forward pass by selectively suppressing specific attention heads.

Let $z_h(x)$ denote the output of attention head $h$ for input $x$. We define a binary gating operator $g_h \in \{0,1\}$ for each head as:
\begin{equation}
g_h =
\begin{cases}
0, & \text{if } h \in \mathcal{H}^{\text{top-}k}, \\
1, & \text{otherwise},
\end{cases}
\end{equation}
and obtain the modified head output:
\begin{equation}
\tilde{z}_h(x) = g_h \cdot z_h(x).
\end{equation}
This yields a perturbed model $f_\theta^{\text{masked}}(x)$.

This operation can be interpreted as a form of \emph{causal editing} on the model’s computation graph, where selected edges (i.e., head outputs) are removed to disrupt specific functional circuits. In particular, masking heads in $\mathcal{H}^{\text{err}}$ suppresses pathways responsible for overconfident errors, while masking heads in $\mathcal{H}^{\text{noise}}$ reduces spurious variation in otherwise correct predictions.

The intervention is both \emph{local} and \emph{minimal}, modifying only a small subset of heads and thus preserving most of the model’s representational capacity. The calibration classes are used exclusively during mask construction. At inference time, the resulting global mask is applied uniformly to all test inputs without requiring ground-truth labels or sample-specific class identification. From a computational perspective, the overhead is negligible, as the intervention involves only element-wise gating during the forward pass. Since no parameters are updated, the method can be seamlessly applied to pretrained models in a plug-and-play manner.

\section{Experiments}

\subsection{Implementation Details}

In practice, we apply {CoCHI} on a small subset of the data (typically fewer than 50 samples), with balanced sampling across calibration classes to ensure stable attribution estimates. We use a fixed confidence threshold $\tau_c = 0.5$ to define calibration classes, and a threshold $\tau$ to identify facilitating heads. The number of masked heads $k$ is treated as a hyperparameter; we find that small values (e.g., $k \in [6,8]$) provide a favourable trade-off between calibration improvement and accuracy preservation. The penalty coefficient was empirically tuned across 15 model--dataset runs, yielding a stable trade-off that limits degradation on \texttt{correct\_high} calibration ($\Delta \mathrm{ECE} < 0.015$) while still effectively suppressing overconfident errors on incorrect predictions. We evaluate our method on three representative open-source large language models: \texttt{LLaMA-3.1-8B}~\cite{grattafiori2024llama}, \texttt{Gemma-2-9B}~\cite{00118}, and \texttt{Mistral-7B-v0.1}~\cite{06825}. We adopt \texttt{LLaMA-3.1-8B} as the primary backbone for ablation and hyperparameter studies. All experiments are conducted in a zero-shot setting, ensuring that improvements in calibration are solely attributable to the proposed inference-time intervention. Further experimental details are provided in Appendix~\ref{sec:appendix_data}.

\subsection{Evaluation Protocols}

\textbf{Datasets.}
We evaluate our method on five widely used multiple-choice question answering benchmarks spanning diverse domains, including general knowledge, science, and medicine, to assess both generalisation and domain-specific robustness of calibration. Measuring Massive Multitask Language Understanding (MMLU)~\cite{hendrycks2021measuring} is a large-scale multi-domain benchmark covering 57 subjects such as STEM, humanities, and social sciences, designed to evaluate general knowledge and reasoning ability. MedMCQA~\cite{pal2022medmcqa} is a large medical QA dataset derived from real-world entrance examinations, focusing on domain-specific knowledge in healthcare. OpenBookQA~\cite{mihaylov2018can} is a science QA benchmark requiring multi-hop reasoning over a small set of core science facts. MedQA-USMLE~\cite{jin2021disease} is a challenging medical benchmark based on USMLE exams, evaluating clinical reasoning and professional knowledge. SciQ~\cite{welbl2017crowdsourcing} is a science QA dataset constructed from educational materials, focusing on factual understanding and basic reasoning. All datasets are formatted as four-choice multiple-choice tasks, and we adopt a unified evaluation protocol. 

\textbf{Baselines.}
We compare our method against several representative calibration approaches, including both post-hoc and inference-time techniques. Temperature Scaling (TS)~\cite{guo2017calibration} is a standard post-hoc method that learns a single scalar temperature $T$ on a held-out validation set to rescale logits. Inference-Time Intervention (ITI)~\cite{li2023inference} modifies model activations during inference using a fixed intervention strength $\alpha$, enabling reliability improvements without retraining. Thermometer (Therm.)~\cite{shen2024thermometer} is a learned calibration method that trains a lightweight MLP-based calibrator on top of model outputs to better align predicted confidence with empirical accuracy. Adaptive Temperature Scaling (ATS)~\cite{xie2024calibrating} extends TS by dynamically adapting the temperature based on input features, allowing more flexible calibration across samples. 

\textbf{Metrics.}
We evaluate performance in terms of both accuracy and calibration. Prediction quality is measured by \textbf{accuracy}, defined as the proportion of correctly predicted answers. Calibration is assessed using \textbf{Expected Calibration Error (ECE)}~\cite{guo2017calibration}, which measures the discrepancy between predicted confidence and empirical accuracy across confidence bins; \textbf{Negative Log-Likelihood (NLL)}, which evaluates the quality of predicted probability distributions by penalising overconfident incorrect predictions; and the \textbf{Brier Score (BS)}, which computes the mean squared error between predicted probabilities and ground-truth labels. Lower values of ECE, NLL, and Brier Score indicate better calibration. Together, these metrics provide a comprehensive assessment of both predictive correctness and the reliability of confidence estimates.

\begin{table}[t]
\centering
\caption{Results on four benchmarks (MMLU, MedMCQA, OpenBookQA, and MedQA-USMLE). Best results are highlighted in bold.}
\setlength{\tabcolsep}{3pt}
\tiny
\begin{tabular}{l |cccc| cccc } 
\toprule

 & \multicolumn{4}{c|}{\textbf{MMLU}} & \multicolumn{4}{c}{\textbf{MedMCQA}} 
  \\
\cmidrule(lr){2-5}\cmidrule(lr){6-9}

\textbf{Method} & Accuracy $\uparrow$ & ECE $\downarrow$ & NLL $\downarrow$ & BS $\downarrow$ 
                & Accuracy $\uparrow$ & ECE $\downarrow$ & NLL $\downarrow$ & BS $\downarrow$ 
                \\
\midrule
\multicolumn{9}{c}{\textbf{Gemma-2-9B }} \\
\midrule
TS 
& 0.6653$\pm$0.0050 & 0.0660$\pm$0.0173 & 0.5060$\pm$0.0335 & 0.1683$\pm$0.0125 
& 0.5607$\pm$0.0186 & 0.0556$\pm$0.0261 & 0.6114$\pm$0.0151 & 0.2122$\pm$0.0066
 \\
 ITI 
& 0.6693$\pm$0.0060 & 0.0605$\pm$0.0062 & 0.5029$\pm$0.0233 & 0.1675$\pm$0.0105 
& 0.5673$\pm$0.0200 & 0.0451$\pm$0.0140 & 0.6085$\pm$0.0154 & 0.2122$\pm$0.0070 \\
Therm.
& 0.6653$\pm$0.0050 & 0.1831$\pm$0.0567 & 1.8729$\pm$1.1455 & 0.2302$\pm$0.0429 
& 0.5607$\pm$0.0186 & 0.2849$\pm$0.0106 & 2.9242$\pm$0.4188 & 0.3293$\pm$0.0165 \\
ATS 
& 0.6653$\pm$0.0050 & 0.3784$\pm$0.0052 & 0.9191$\pm$0.0066 & 0.3553$\pm$0.0030 
& 0.5607$\pm$0.0186 & 0.3534$\pm$0.0783 & 4.1559$\pm$1.3606 & 0.3702$\pm$0.0396  \\
CoCHI 
& \textbf{0.6730$\pm$0.0023} & \textbf{0.0601$\pm$0.0011} & \textbf{0.4982$\pm$0.0141} & \textbf{0.1654$\pm$0.0056} 
& \textbf{0.5873$\pm$0.0023} & \textbf{0.0380$\pm$0.0263} & \textbf{0.5983$\pm$0.0064} & \textbf{0.2070$\pm$0.0027} \\
\midrule
\multicolumn{9}{c}{\textbf{LLaMA-3.1-8B}} \\
\midrule
TS 
& 0.6007$\pm$0.0061 & 0.0569$\pm$0.0040 & 0.5205$\pm$0.0138 & 0.1936$\pm$0.0065 
& 0.5753$\pm$0.0297 & 0.0557$\pm$0.0124 & 0.6173$\pm$0.0142 & 0.2144$\pm$0.0052 \\
ITI 
& 0.5853$\pm$0.0685 & 0.2014$\pm$0.1191 & 0.7161$\pm$0.1590 & 0.2580$\pm$0.0643 
& 0.3747$\pm$0.0893 & 0.3766$\pm$0.2287 & 1.2043$\pm$0.7682 & 0.3831$\pm$0.1870 \\
Therm. 
& 0.6007$\pm$0.0061 & 0.1717$\pm$0.0778 & 1.9024$\pm$0.7247 & 0.2415$\pm$0.0354 
& 0.5753$\pm$0.0297 & 0.2327$\pm$0.0356 & 1.8116$\pm$0.8973 & 0.2907$\pm$0.0208 \\
ATS 
& 0.6007$\pm$0.0061 & 0.3265$\pm$0.0257 & 1.7325$\pm$1.5251 & 0.3330$\pm$0.0157 
& 0.5753$\pm$0.0301 & 0.3307$\pm$0.0727 & 2.5476$\pm$2.9442 & 0.3495$\pm$0.0501 \\
CoCHI
& \textbf{0.6120$\pm$0.0626} & \textbf{0.0490$\pm$0.0089} & \textbf{0.5168$\pm$0.0236} & \textbf{0.1743 $\pm$0.0008} 
& \textbf{0.5773$\pm$0.0081} & \textbf{0.0365$\pm$0.0090} & \textbf{0.6064$\pm$0.0032} & \textbf{0.2114$\pm$0.0012} \\
            \midrule
\multicolumn{9}{c}{\textbf{Mistral-7B}} \\
\midrule
TS 
& 0.5987$\pm$0.0121 & 0.0902$\pm$0.0182 & 0.5754$\pm$0.0130 & 0.1974$\pm$0.0037 
& \textbf{0.4867$\pm$0.0081} & 0.0604$\pm$0.0308 & 0.6424$\pm$0.0339 & 0.2249$\pm$0.0133 \\
ITI 
& 0.3980$\pm$0.1152 & 0.3693$\pm$0.2666 & 1.2168$\pm$0.8397 & 0.3845$\pm$0.2056 
& 0.3293$\pm$0.0939 & 0.3985$\pm$0.3280 & 1.4918$\pm$1.2039 & 0.4317$\pm$0.2469 \\
Therm. 
& 0.5987$\pm$0.0121 & 0.2988$\pm$0.0449 & 1.4277$\pm$0.9584 & 0.3177$\pm$0.0274 
& \textbf{0.4867$\pm$0.0081} & 0.2668$\pm$0.0776 & 2.6632$\pm$0.6919 & 0.3385$\pm$0.0372 \\
ATS 
& 0.5987$\pm$0.0121 & 0.3379$\pm$0.0290 & 2.5765$\pm$1.4883 & 0.3462$\pm$0.0184 
& \textbf{0.4867$\pm$0.0081} & 0.3817$\pm$0.1376 & 3.6701$\pm$2.4950 & 0.4032$\pm$0.0945 \\
CoCHI
& \textbf{0.6100$\pm$0.0040} & \textbf{0.0854$\pm$0.0015} & \textbf{0.5649$\pm$0.0008} & \textbf{0.1941$\pm$0.0005} 
& 0.4847$\pm$0.0023 & \textbf{0.0383$\pm$0.0035} & \textbf{0.6168$\pm$0.0012} & \textbf{0.2146$\pm$0.0006} 
 \\
\bottomrule
\end{tabular}
\\[0.5em]
\begin{tabular}{l |cccc |cccc } 
\toprule

 & \multicolumn{4}{c|}{\textbf{OpenBookQA}} & \multicolumn{4}{c}{\textbf{MedQA-USMLE}} 
  \\
\cmidrule(lr){2-5}\cmidrule(lr){6-9}

\textbf{Method} & Accuracy $\uparrow$ & ECE $\downarrow$ & NLL $\downarrow$ & BS $\downarrow$ 
                & Accuracy $\uparrow$ & ECE $\downarrow$ & NLL $\downarrow$ & BS $\downarrow$ 
                \\
\midrule
\multicolumn{9}{c}{\textbf{Gemma-2-9B }} \\
\midrule
TS 
& 0.6080$\pm$0.0087 & 0.0686$\pm$0.0119 & 0.5410$\pm$0.0076 & 0.1848$\pm$0.0054 
& 0.7714$\pm$0.0124 & 0.1782$\pm$0.0188 & 0.5109$\pm$0.0377 & 0.1829$\pm$0.0118 
 \\
 ITI 
& \textbf{0.6100$\pm$0.0111} & 0.0682$\pm$0.0064 & 0.5438$\pm$0.0070 & 0.1848$\pm$0.0043 
& \textbf{0.7927$\pm$0.0151} & 0.1725$\pm$0.0143 & 0.5091$\pm$0.0163 & 0.1708$\pm$0.0072 \\
Therm.
& 0.6080$\pm$0.0087 & 0.2635$\pm$0.0266 & 2.4829$\pm$0.5178 & 0.2930$\pm$0.0223 
& 0.7810$\pm$0.0114 & 0.1767$\pm$0.0298 & 1.7598$\pm$0.2096 & 0.1757$\pm$0.0205  \\
ATS 
& 0.6080$\pm$0.0087 & 0.3350$\pm$0.0165 & 1.9401$\pm$1.8405 & 0.3399$\pm$0.0090 
& 0.7740$\pm$0.0164 & 0.1863$\pm$0.0131 & 2.2281$\pm$0.2476 & 0.1866$\pm$0.0141   \\
CoCHI 
& 0.6090$\pm$0.0766 & \textbf{0.0434$\pm$0.0116} & \textbf{0.5353$\pm$0.0268} & \textbf{0.1840$\pm$0.0092} 
& 0.7893$\pm$0.0223 & \textbf{0.1723$\pm$0.0070} & \textbf{0.5081$\pm$0.0046} & \textbf{0.1694$\pm$0.0019}  \\
\midrule
\multicolumn{9}{c}{\textbf{LLaMA-3.1-8B}} \\
\midrule
TS 
& \textbf{0.6827$\pm$0.0170} & 0.0618$\pm$0.0106 & 0.5776$\pm$0.0125 & 0.1853$\pm$0.0066 
& \textbf{0.7360$\pm$0.0087} & 0.1198$\pm$0.0200 & 0.5603$\pm$0.0394 & 0.1750$\pm$0.0123  \\
ITI 
& 0.5200$\pm$0.1618 & 0.2272$\pm$0.2848 & 0.9131$\pm$0.6744 & 0.2952$\pm$0.2009 
& 0.5387$\pm$0.0908 & 0.1611$\pm$0.0761 & 0.6333$\pm$0.0739 & 0.2226$\pm$0.0332  \\
Therm. 
& \textbf{0.6827$\pm$0.0170} & 0.1654$\pm$0.0405 & 1.5188$\pm$0.5910 & 0.2180$\pm$0.0302 
& \textbf{0.7360$\pm$0.0087} & 0.1954$\pm$0.0372 & 1.9680$\pm$0.4174 & 0.2124$\pm$0.0213  \\
ATS 
& \textbf{0.6827$\pm$0.0170} & 0.3346$\pm$0.0658 & 2.6951$\pm$1.5194 & 0.3195$\pm$0.0403 
& \textbf{0.7360$\pm$0.0087} & 0.2432$\pm$0.0119 & 3.1290$\pm$0.1205 & 0.2430$\pm$0.0082  \\
CoCHI
& 0.6700$\pm$0.0040 & \textbf{0.0523$\pm$0.0193} & \textbf{0.5069$\pm$0.0090} & \textbf{0.1730$\pm$0.0035} 
& 0.7307$\pm$0.0186 & \textbf{0.0961$\pm$0.0169} & \textbf{0.5066$\pm$0.0121} & \textbf{0.1700$\pm$0.0048}  \\
            \midrule
\multicolumn{9}{c}{\textbf{Mistral-7B}} \\
\midrule
TS 
& 0.5100$\pm$0.0100 & 0.0673$\pm$0.0383 & 0.6016$\pm$0.0130 & 0.2076$\pm$0.0033 
& \textbf{0.6707$\pm$0.0064} & 0.1489$\pm$0.0077 & 0.5720$\pm$0.0115 & \textbf{0.1987$\pm$0.0089}  \\
ITI 
& 0.3567$\pm$0.0996 & 0.3583$\pm$0.2315 & 1.0117$\pm$0.3433 & 0.3684$\pm$0.1309 
& 0.5193$\pm$0.0574 & 0.1522$\pm$0.0597 & 0.6011$\pm$0.0402 & 0.2089$\pm$0.0195  \\
Therm. 
& 0.5100$\pm$0.0100 & 0.2388$\pm$0.0647 & 2.0444$\pm$0.3463 & 0.3042$\pm$0.0360 
& \textbf{0.6707$\pm$0.0064} & 0.2214$\pm$0.0398 & 2.0876$\pm$0.3632 & 0.2511$\pm$0.0222  \\
ATS 
& 0.5100$\pm$0.0100 & 0.4317$\pm$0.0040 & 4.2157$\pm$0.0519 & 0.4266$\pm$0.0012 
& \textbf{0.6707$\pm$0.0064} & 0.2904$\pm$0.0064 & 3.0443$\pm$0.0280 & 0.2896$\pm$0.0046 \\
CoCHI
& \textbf{0.5120$\pm$0.0040} & \textbf{0.0316$\pm$0.0073} & \textbf{0.5912$\pm$0.0029} & \textbf{0.2047$\pm$0.0014} 
& 0.6700$\pm$0.0139 & \textbf{0.1432$\pm$0.0108} & \textbf{0.5634$\pm$0.0068} & 0.2028$\pm$0.0035 
 \\
\bottomrule
\end{tabular}
\label{tab:main}
\end{table}

\subsection{Main Results}

The results in Table~\ref{tab:main} show that \textbf{CoCHI} consistently improves calibration across all datasets and models while maintaining competitive accuracy. On datasets such as \textit{MMLU} and \textit{MedMCQA}, CoCHI achieves clear reductions in ECE together with improved NLL and Brier Score. Similar trends are observed on \textit{OpenBookQA} and \textit{MedQA-USMLE}, where CoCHI attains the lowest or near-lowest calibration errors without sacrificing predictive performance. In several cases, CoCHI also improves accuracy, particularly on LLaMA-3.1-8B and Mistral-7B.

In contrast, existing methods show limited or unstable behaviour across datasets. Post-hoc methods such as TS and ATS provide only marginal improvements and can severely degrade calibration in some settings, while Thermometer often increases NLL, indicating unreliable probability estimates. ITI also exhibits high variance across models and datasets. The advantage of CoCHI is especially evident on Mistral-7B, where it consistently achieves lower ECE together with strong NLL and Brier Score performance. Overall, these results suggest that intervening on targeted causal attention heads provides more effective and stable calibration than conventional output-level adjustment methods.
 Additional results are provided in Appendix~\ref{Main}.

\subsection{Ablation Study}

\begin{wraptable}{r}{0.55\textwidth}
\vspace{-0.5cm}
    \centering
    \renewcommand\arraystretch{1} 
    \setlength{\tabcolsep}{4pt} 
    \caption{Ablation study of different masking strategies on MMLU. }
    \label{tab:mask_ablation}
    \vspace{5pt}
    \scriptsize
    \begin{tabular}{lcc}
    \toprule
    \textbf{Mask Rule} & \textbf{Masked Acc $\uparrow$} & \textbf{Masked ECE $\downarrow$} \\
    \midrule
    Wrong-high facilitating only & 0.2044$\pm$0.0038 & 0.4745$\pm$0.1477 \\
    Correct-high non-facilitating only & 0.2389$\pm$0.0398 & 0.1518$\pm$0.0474 \\
    Correct-high interfering only & 0.4267$\pm$0.0462 & 0.0694$\pm$0.0488 \\
    Canonical union (broad) & 0.2189$\pm$0.0271 & 0.3850$\pm$0.1816 \\
    Random size-matched & 0.2111$\pm$0.0077 & 0.4330$\pm$0.2736 \\
    \midrule
    {CoCHI} & \textbf{0.6120$\pm$0.0626} & \textbf{0.0490$\pm$0.0089} \\
    \bottomrule
    \end{tabular}
\end{wraptable}

The ablation results in Table~\ref{tab:mask_ablation} highlight the importance of structured, causally-informed masking. Naive strategies, such as masking only wrong-high facilitating heads or applying a broad canonical-union mask, severely degrade both accuracy and calibration (accuracy 0.2044/0.2189; ECE 0.4745/0.3850), suggesting that indiscriminate head removal disrupts essential model functionality. More targeted variants also fail to achieve a satisfactory trade-off: masking correct-high interfering heads lowers ECE to 0.0694 but substantially reduces accuracy to 0.4267, while random size-matched masking performs poorly (accuracy 0.2111, ECE 0.4330). In contrast, \textbf{CoCHI} achieves the best balance, with accuracy 0.6120 and ECE 0.0490. These results show that effective intervention depends not only on identifying problematic heads, but also on selecting a compact set of high-impact components through canonical-union design and top-$k$ ranking.

\subsection{Sensitivity Analysis}
\begin{wraptable}{r}{0.6\textwidth}
\vspace{-0.5cm}
	\centering
	\caption{Sensitivity analysis of the confidence threshold $\tau_c$ on MMLU.}
	\label{tab:conf_threshold}
	\renewcommand\arraystretch{1} 
    \setlength{\tabcolsep}{4pt} 
    \vspace{5pt}
    \scriptsize
    \begin{tabular}{lcccc}
    \toprule
    \textbf{$\tau_c$} & \textbf{Accuracy $\uparrow$} & \textbf{ ECE $\downarrow$} & \textbf{ NLL $\downarrow$} & \textbf{ BS $\downarrow$} \\
    \midrule
     0.2 & 0.5933 $\pm$0.0217 & 0.0555 $\pm$0.0096 & 0.5385 $\pm$0.0075 & 0.1808 $\pm$0.0032 \\
 0.3 & 0.6060 $\pm$0.0059 & 0.0643 $\pm$0.0102 & 0.5319 $\pm$0.0043 & 0.1781 $\pm$0.0017 \\
 0.4 & 0.5860 $\pm$0.0118 & 0.0665 $\pm$0.0108 & 0.5276 $\pm$0.0058 & 0.1761 $\pm$0.0025 \\
 0.5 & 0.6120$\pm$0.0626 & 0.0490$\pm$0.0089 & 0.5168$\pm$0.0236 & 0.1743 $\pm$0.0008 \\
 0.7 & 0.5873 $\pm$0.0377 & 0.0705 $\pm$0.0099 & 0.5548 $\pm$0.0212 & 0.1873 $\pm$0.0083 \\
 0.9 & 0.5933 $\pm$0.0081 & 0.0696 $\pm$0.0277 & 0.5255 $\pm$0.0059 & 0.1748 $\pm$0.0022 \\
    \bottomrule
    \end{tabular}
    \end{wraptable}
    
Table~\ref{tab:conf_threshold} presents the sensitivity analysis of the confidence threshold $\tau_c$. Overall, the proposed method remains relatively stable across different threshold settings, indicating that its performance is not highly sensitive to the choice of $\tau_c$. Among all configurations, $\tau_c=0.5$ achieves the best overall trade-off, yielding the lowest ECE, while also obtaining competitive Accuracy, NLL, and Brier Score values. In contrast, very high thresholds (e.g., $\tau_c=0.7$ and $0.9$) lead to noticeable degradation in calibration quality, reflected by increased ECE and NLL. Lower thresholds such as $0.2$ provide reasonable calibration performance but slightly reduce predictive accuracy. These results suggest that moderate confidence thresholds are more effective for balancing classification performance and uncertainty calibration.

\subsection{Efficiency Analysis}
\begin{wraptable}{r}{0.65\textwidth}
\vspace{-0.5cm}
	\centering
    \caption{Efficiency comparison of calibration methods using 40 calibration samples on MMLU.}
	\label{tab:efficiency_main}
	\renewcommand\arraystretch{1} 
    \setlength{\tabcolsep}{5pt} 
    \vspace{5pt}
    \scriptsize
    \begin{tabular}{lccccc}
    \toprule
    \textbf{Method} & \textbf{GPU} & \textbf{Cal. time (s) $\downarrow$} & \textbf{Artifact size $\downarrow$} & \textbf{Inf.} & \textbf{Prunable} \\
    \midrule
    TS & No & 0.5 & 4 B & Negligible & No \\
    ATS & No & 25.8 & 16 B & Negligible & No \\
    ITI & Yes & 18.4 & 16 KB & $O(k \cdot d)$  & No \\
    Therm. & No & 19.4 & 1.0 MB & $O(H \cdot C)$  & No \\\midrule
    CoCHI & Yes & \textbf{69.9} & \textbf{40 B} & \textbf{None} & \textbf{Yes} \\
    \bottomrule
    \end{tabular}
\end{wraptable}
Table~\ref{tab:efficiency_main} presents an efficiency comparison across calibration methods, highlighting a clear trade-off between calibration cost and inference overhead. {CoCHI} requires GPU computation during a one-time calibration phase (69.9s), but produces an extremely compact artifact (40 B) and introduces zero inference overhead, as the intervention is implemented by directly zeroing selected attention heads. In contrast, methods such as ITI and Thermometer incur additional computation at every forward pass (e.g., $O(k \cdot d)$ and $O(H \cdot C)$, respectively), leading to ongoing runtime costs and significantly larger artifacts (up to 1.0 MB). Lightweight post-hoc approaches such as TS and ATS have minimal calibration cost (0.5s and 25.8s) and negligible inference overhead, but do not modify the model’s internal behaviour and therefore lack structural impact.  

A key advantage of CoCHI is its \emph{prunability}: the masked heads can be permanently removed, yielding a smaller model with no runtime wrapper. Overall, CoCHI shifts the computational burden from continuous inference to a one-time structural calibration step, offering a more deployment-efficient solution for improving model reliability. More results are available in Appendix~\ref{Efficiency}.

\subsection{Case Study}
\begin{wrapfigure}{r}{0.6\textwidth} 
\vspace{-0.5cm}
    \centering
    \includegraphics[width=0.6\textwidth]{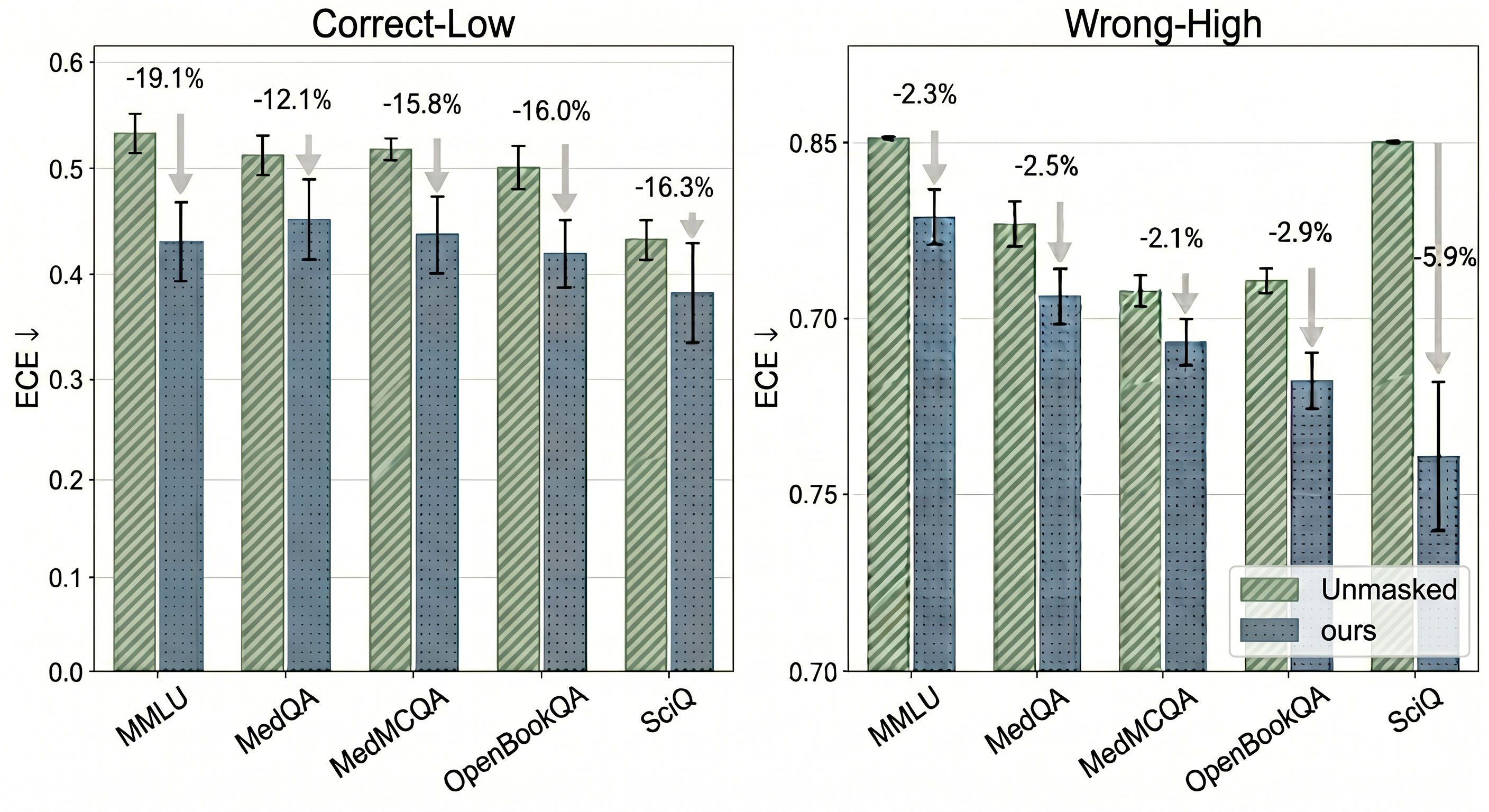}
    \vspace{-0.25cm}
    \caption{ECE comparison between the unmasked model and CoCHI on \texttt{correct\_low} and \texttt{wrong\_high} subsets.}
    \label{fig:case}
\end{wrapfigure}
Figure~\ref{fig:case} presents a per-class calibration comparison between the unmasked baseline and our method across five datasets, focusing on correct predictions with low confidence (left) and wrong predictions with high confidence (right). While our method is primarily designed to address \texttt{wrong\_high} errors, we observe consistent improvements in both regimes. In the \texttt{correct\_low} subset, our method reduces ECE across all datasets (e.g., 0.10 on MMLU and 0.09 on MedMCQA), indicating improved confidence alignment for underconfident correct predictions. In the more challenging \texttt{wrong\_high} subset, our method also consistently lowers ECE, albeit with smaller margins. This suggests that suppressing causally relevant heads not only mitigates overconfident errors but also improves overall confidence stability.

\section{Conclusion}

In this work, we study overconfident errors in large language models and argue that miscalibration should be addressed not only at the output level, but also through the internal mechanisms that give rise to unreliable confidence estimates. We propose {CoCHI}, a class-oriented causal head intervention framework that decomposes predictions by correctness and confidence, identifies attention heads associated with distinct calibration behaviours, and performs targeted inference-time masking without updating model parameters. Across multiple LLMs and five multiple-choice QA benchmarks, CoCHI consistently improves calibration metrics while maintaining competitive predictive accuracy. The ablation and sensitivity analyses further show that reliable calibration depends on structured causal selection rather than indiscriminate head removal. Moreover, while CoCHI primarily targets \texttt{wrong\_high} errors, it also yields improvements in \texttt{correct\_low} cases. These findings support a mechanism-aware view of calibration, linking model reliability to identifiable internal structures rather than treating it solely as a post-hoc output adjustment problem.

\paragraph{Limitations.}
This work focuses on multiple-choice question answering, where confidence can be naturally defined over a fixed answer space. Extending CoCHI to open-ended generation, long-form reasoning, and multi-turn interaction requires further investigation, especially when correctness and confidence are harder to define. In addition, our current intervention operates at the attention-head level, which provides an efficient and interpretable unit for masking, but may not capture all sources of miscalibration arising from MLP layers, residual streams, or cross-layer interactions.

\section*{Acknowledgements}
We acknowledge the RMIT Advanced Compute Environment (RACE) for providing the high-performance computing resources and infrastructure that supported this research. We also thank Patrick Taylor for his valuable technical support and assistance.

\bibliographystyle{unsrtnat}
\bibliography{RiskMask}


\appendix

\section{Dataset Preparation and Experimental Pipeline}
\label{sec:appendix_data}

This section details the data preparation and experimental pipeline to ensure reproducibility.

\subsection{Datasets}

We use five publicly available multiple-choice question answering datasets (as shown in Table~\ref{Datasets}) from HuggingFace, covering general knowledge, science, and medical domains. All datasets are formatted as four-option multiple-choice questions with a single correct answer.

\begin{table}[h]
\centering
\small
\caption{Statistical summary of the datasets.}
\begin{tabular}{lcc}
\toprule
\textbf{Dataset} & \textbf{Domain} & \textbf{\# Questions} \\
\midrule
MMLU & Multi-domain knowledge & 14,042 \\
MedQA-USMLE & Medical exam & 11,451 \\
OpenBookQA & Science & $\sim$4,200 \\
SciQ & Science & 13,679 \\
MedMCQA & Medical exam & 187,005 \\
\bottomrule
\end{tabular}\label{Datasets}
\end{table}

\subsection{Data Cleaning and Standardisation}

All datasets are converted into a unified format containing: question ID, dataset source, question text, four answer choices (A--D), and the correct label. We remove samples that: (i) do not contain exactly four options, or (ii) have missing or invalid answer labels.

\paragraph{SciQ Adjustment.}
In SciQ, the correct answer appears in a fixed position. To avoid positional bias, we cyclically rotate answer options to ensure a uniform distribution.

\subsection{Prompt Formatting}

All inputs follow a unified zero-shot prompt:

\begin{quote}
Answer the following multiple choice question with a single letter: A, B, C, or D.

Question: [question text]

A. [choice 1] \\
B. [choice 2] \\
C. [choice 3] \\
D. [choice 4]

Answer:
\end{quote}

No demonstrations or additional context are provided.











\subsection{Training Example Selection}

From the fit pool, we sample a balanced calibration subset:
\begin{itemize}
    \item up to 10 examples per class,
    \item maximum of 40 samples per dataset.
\end{itemize}

This simulates a low-resource calibration setting.

\subsection{Top-$k$ Head Selection}

We determine the number of masked heads $k$ via grid search over $k \in \{1, \dots, 20\}$. Selection is based on minimising Expected Calibration Error (ECE), subject to an accuracy drop constraint of at most 1\% relative to the unmodified model.
A single $k$ is selected per model and applied across all datasets to capture model-level calibration properties. The selected values are $k=10$ for LLaMA-3.1-8B, and $k=8$ for Gemma-2-9B and Mistral-7B.

\begin{table}[h]
\centering
\caption{Results on SciQ. Best results are highlighted in bold.}
\small
\setlength{\tabcolsep}{3pt}
\begin{tabular}{l cccc } 
\toprule

\textbf{Method} & Acc$\uparrow$ & ECE$\downarrow$ & NLL$\downarrow$ & BS$\downarrow$ 
                \\
\midrule
\multicolumn{5}{c}{\textbf{Gemma-2-9B }} \\
\midrule
TS 
& 0.9173$\pm$0.0012 & 0.0579$\pm$0.0184 & 0.4699$\pm$0.4226 & 0.0641$\pm$0.0094 \\
 ITI 
& 0.9253$\pm$0.0083 & 0.1027$\pm$0.0091 & 0.2337$\pm$0.0076 & 0.0617$\pm$0.0025 \\
Therm.
& 0.9223$\pm$0.0021 & 0.0609$\pm$0.0066 & 1.0836$\pm$0.1824 & 0.0599$\pm$0.0068 \\
ATS 
& 0.9173$\pm$0.0012 & 0.0607$\pm$0.0067 & 1.1578$\pm$0.1183 & 0.0602$\pm$0.0065 \\
CoCHI 
& \textbf{0.9353$\pm$0.1069} & \textbf{0.0563$\pm$0.0199} & \textbf{0.2309$\pm$0.1276} & \textbf{0.0548$\pm$0.0472} \\
\midrule
\multicolumn{5}{c}{\textbf{LLaMA-3.1-8B}} \\
\midrule
TS 
& 0.8790$\pm$0.0210 & 0.0354$\pm$0.0109 & 0.2377$\pm$0.0145 & 0.0721$\pm$0.0045 \\
ITI 
& 0.8647$\pm$0.0311 & 0.0359$\pm$0.0176 & 0.3032$\pm$0.0329 & 0.0921$\pm$0.0139 \\
Therm. 
& 0.8920$\pm$0.0131 & 0.0738$\pm$0.0112 & 0.9514$\pm$0.2711 & 0.0807$\pm$0.0065 \\
ATS 
& 0.8790$\pm$0.0210 & 0.0932$\pm$0.0219 & 1.4970$\pm$0.2712 & 0.0928$\pm$0.0195 \\
CoCHI
& \textbf{0.9020$\pm$0.0020} & \textbf{0.0330$\pm$0.0010} & \textbf{0.2266$\pm$0.0012} & \textbf{0.0676$\pm$0.0004} \\
            \midrule
\multicolumn{5}{c}{\textbf{Mistral-7B}} \\
\midrule
TS 
& 0.8667$\pm$0.0150 & 0.0764$\pm$0.0048 & 0.3057$\pm$0.0402 & 0.0944$\pm$0.0045 \\

ITI 
& 0.5720$\pm$0.1512 & 0.2095$\pm$0.1429 & 0.6363$\pm$0.1672 & 0.2249$\pm$0.0680 \\
Therm. 
& 0.8467$\pm$0.0310 & 0.0986$\pm$0.0084 & 1.4162$\pm$0.0701 & 0.0978$\pm$0.0073 \\
ATS 
& 0.8667$\pm$0.0150 & 0.1056$\pm$0.0139 & 1.6920$\pm$0.2413 & 0.1049$\pm$0.0134 \\
CoCHI
& \textbf{0.8773$\pm$0.0046} & \textbf{0.0730$\pm$0.0116} & \textbf{0.3030$\pm$0.0144} & \textbf{0.0903$\pm$0.0046} \\
\bottomrule
\end{tabular}
\label{tab:main3}
\end{table}





\section{Additional Results}
\subsection{Main results}
\label{Main}
Table~\ref{tab:main3} confirms the effectiveness of our method on SciQ, demonstrating consistent improvements in calibration across all models while maintaining strong predictive performance. On {Gemma-2-9B}, our method achieves the best overall calibration, reducing ECE and improving both NLL and Brier Score , while matching the highest accuracy. On {LLaMA-3.1-8B}, the gains are even more pronounced: our method simultaneously achieves the best performance across all four metrics, improving accuracy, lowering ECE, and reducing both NLL and Brier Score. Similar trends are observed on {Mistral-7B}, where our method achieves the best results across all metrics, outperforming the strongest baseline (ATS) in both calibration and accuracy. These results indicate that the proposed intervention consistently improves confidence alignment without sacrificing predictive quality.

In contrast, baseline methods exhibit unstable and inconsistent behaviour. ITI achieves competitive accuracy on Gemma-2-9B but significantly worsens calibration, while on Mistral-7B it severely degrades both accuracy and calibration. Post-hoc methods such as TS and ATS provide moderate or inconsistent calibration improvements but often fail to jointly optimise NLL and Brier Score, with ATS in particular showing substantial degradation in probabilistic metrics (e.g., NLL exceeding 1.0 across models). Thermometer similarly leads to large increases in NLL, indicating unreliable probability estimates. Overall, these results reinforce our central claim: calibration is better improved by targeting internal causal mechanisms rather than relying on output-level adjustments, leading to more stable and reliable performance across diverse models.

\subsection{Efficiency Analysis}
\label{Efficiency}
Table~\ref{tab:efficiency_models} shows that the calibration cost of our method scales predictably with model size while remaining practical. The average calibration time increases from 39\,s for Mistral-7B to 58\,s for LLaMA-3.1-8B and 88\,s for Gemma-2-9B, reflecting the expected growth in forward-pass complexity, yet all models complete calibration within roughly 1–1.5 minutes. The variance is moderate (std $\approx$ 20–23\,s), indicating stable computational behaviour. Peak GPU memory ranges from 28.1\,GiB to 46.2\,GiB, increasing with model size but remaining within the capacity of standard high-end GPUs. Notably, these costs include multiple class-conditioned forward passes, which are essential to our method, yet remain comparable to inference-time intervention baselines. Overall, the results indicate that the calibration procedure introduces only moderate overhead, scales with model size, and is practical for real-world deployment.

\begin{table}[t]
\centering
\small
\caption{
Calibration time (s; mean $\pm$ std) of CoCHI across three model families, averaged over 15 runs per model. Costs include class-conditioned forward passes and exclude model loading, and scale with model size while remaining practical.
}
\begin{tabular}{lcccc}
\toprule
\textbf{Model} & \textbf{Cal. time (s) $\downarrow$} & \textbf{Min} & \textbf{Max} & \textbf{Peak GPU (GiB)} \\
\midrule
Mistral-7B-v0.1 & 39$\pm$21 & 21 & 80 & 28.1 \\
Llama-3.1-8B & 58$\pm$20 & 41 & 93 & 30.4 \\
Gemma-2-9B & 88$\pm$23 & 72 & 159 & 46.2 \\
\bottomrule
\end{tabular}

\label{tab:efficiency_models}
\end{table}


\end{document}